\documentclass[runningheads,a4paper]{llncs}
\usepackage[T1]{fontenc}
\usepackage{amsmath}
\usepackage[markup=default]{changes}
\usepackage{placeins}  
\usepackage[misc,geometry]{ifsym}
\usepackage{multicol}
\usepackage{rotating}
\usepackage[hidelinks]{hyperref}
\usepackage{url}
\usepackage{makecell} 
\usepackage{multirow}
\usepackage{xcolor}
\usepackage{graphicx,verbatim}
\usepackage{booktabs}
\usepackage{amssymb}
\usepackage{tabularx,array}
\usepackage[normalem]{ulem}
\usepackage{xurl}
\newcommand{\rev}[1]{#1}  
\begin{document}
%
\title{Vis2Reg: Visibility-Aware Landmark-Free Geometric 3D--2D Registration for Liver Laparoscopy}
\titlerunning{Vis2Reg: Visibility-Aware 3D--2D Liver Registration}
%
%
%

\author{Jiaming Feng\inst{1}\orcidID{0009-0009-0506-7152} \and
Xukun Zhang\inst{2}\orcidID{0000-0003-2869-9434} \and
Shahid Farid\inst{3}\orcidID{0000-0001-7685-182X}\and
Sharib Ali(\Letter) \inst{1} \orcidID{0000-0003-1313-3542}}

\authorrunning{J. Feng et al.}
\institute{AI in Medicine and Surgery Group, School of Computer Science, University of Leeds, LS2 9JT, Leeds, United Kingdom\\
\email{\{vqkc0507, s.s.ali\}@leeds.ac.uk}
\and
Department of Diagnostic Radiology, Li Ka Shing Faculty of Medicine, The University of Hong Kong, Pok Fu Lam, Hong Kong\\
\email{xukunzhg@hku.hk}
\and
Department of HPB and Transplant Surgery, St.~James's University Hospital, Leeds, United Kingdom\\
\email{s.farid@nhs.net}\\[2pt]
}
    
\maketitle              

\begin{abstract}
Accurate 3D--2D liver registration, which aligns preoperative 3D models to partial, view-dependent intraoperative surface observations, is critical for AR-guided laparoscopic surgery but remains challenging due to severe occlusion, limited visibility, and the lack of 3D ground-truth supervision. Existing landmark-free approaches perform partial-to-complete geometric alignment, yet robust self-supervision under extreme partial visibility remains difficult.
We propose Vis2Reg, a visibility-aware registration framework that explicitly constrains deformation using mask-consistent visible regions. We introduce a visibility-aware self-supervision that derives a visible-domain 3D supervision signal from intraoperative masks, enabled by differentiable point rasterization and mask-guided back-projection. This formulation improves robustness under severe occlusion while maintaining fully self-supervised learning. Vis2Reg combines a robust geometric rigid initialization module with an implicit neural deformation field for stable alignment. Vis2Reg achieves a Dice score of 92.6\% and a Chamfer Distance of 1.43 mm on real intraoperative datasets, with 111 ms per-frame inference time, demonstrating both accuracy and practical efficiency.

\end{abstract}



\section{Introduction}
Minimally invasive laparoscopic and robotic liver resection is increasingly adopted, but surgeons rely on a narrow field-of-view with no direct access to sub-surface anatomy, limiting assessment of tumour--vessel relationships and increasing surgical risk~\cite{feuerstein2008intraoperative,ali2025objective}.
Augmented reality (AR) can mitigate this by overlaying patient-specific preoperative 3D liver, vascular, or tumour models onto the intraoperative video, improving spatial awareness beyond the visible surface~\cite{ali2025objective,prevost2020efficiency,ramalhinho2023value}. Accurate, robust preoperative-to-intraoperative fusion of the deformable liver is therefore critical for reliable AR guidance.

At its core, AR-guided laparoscopy requires preoperative-to-intraoperative 3D--2D registration of a preoperative liver model to intraoperative views and camera geometry, which is ill-posed under large non-rigid deformation, adverse imaging (occlusion, specularities, smoke, blood), and partial, view-dependent observations~\cite{espinel2022multiview,mhiri2025neural,robu2018global,koo2022automatic,maierhein2013optical}. Early approaches relied on landmarks, contours, or biomechanical constraints~\cite{espinel2022multiview,besl1992icp,neri2025surgical_ar_review}, but are brittle under occlusion and low overlap~\cite{maierhein2013optical,robu2018global,neri2025surgical_ar_review}. \rev{This family spans classical iterative closest point-based (ICP) optimization, biomechanical finite element method (FEM) simulation, and shape-matching such as functional maps~\cite{besl1992icp,ozgur2018biomechanical,ovsjanikov2012functional}.} Learning-based approaches predict deformation directly from observations~\cite{mhiri2025neural} yet remain limited by scarce supervision and incomplete views~\cite{ali2025objective,neri2025surgical_ar_review}.

A recent paradigm shift reformulates laparoscopic registration as \emph{partial-to-complete 3D registration}, aligning sparse intraoperative surface observations with the full preoperative liver geometry \cite{zhou_landmark-free_2025,huang2025landmarkfree,neri2025benchmarking_partial}.
This formulation enables self-supervised learning using readily available signals such as segmentation masks and geometric consistency, eliminating the need for explicit 3D correspondence supervision \cite{zhou_landmark-free_2025,huang2025landmarkfree}.
Nevertheless, intraoperative supervision is fundamentally visibility limited: only the mask-consistent visible surface provides reliable geometric constraints, and explicitly modeling such visibility-consistent supervision to guide deformation remains challenging under severe occlusion and limited overlap \cite{guan2023intraoperative,dai2025deep_graph_matching,robu2018global,maierhein2013optical,neri2025benchmarking_partial}.
Furthermore, rigid initialization is often unstable when intraoperative observations are sparse and noisy \cite{robu2018global,koo2022automatic,huang2025landmarkfree,neri2025benchmarking_partial}. \rev{Generic low-overlap point-cloud registration methods (e.g., PREDATOR~\cite{huang2021predator}, GeoTransformer~\cite{qin_geotransformer_2023}) improve correspondence robustness but are not tailored to visibility-limited surgical self-supervision.}
To address these challenges, we propose \textit{Vis2Reg}, a visibility-aware, landmark-free, geometry-only framework for self-supervised laparoscopic liver registration under severe partial visibility.
It derives mask-consistent visible-domain 3D supervision from intraoperative masks via differentiable point rasterization and mask-guided back-projection \cite{ravi2020pytorch3d}, focusing learning on truly observable geometry.
This improves deformation robustness while preserving anatomical plausibility, and enables geometry-only inference without masks or explicit correspondences.

Our contributions are three-fold:
(1) We introduce a visibility-aware self-supervision formulation that constructs mask-consistent visible-domain 3D supervision signals for landmark-free laparoscopic liver registration;
(2) we develop a differentiable point rasterization and mask-guided back-projection mechanism to derive visibility-consistent geometric supervision from intraoperative observations;
and (3) we integrate robust geometric rigid initialization with an implicit neural deformation field, achieving improved registration robustness and near-real-time performance on in-vivo laparoscopic datasets.

\section{Method}

\subsection{Problem formulation}
\begin{figure}[!t]
    \centering
    \includegraphics[width=1\linewidth]{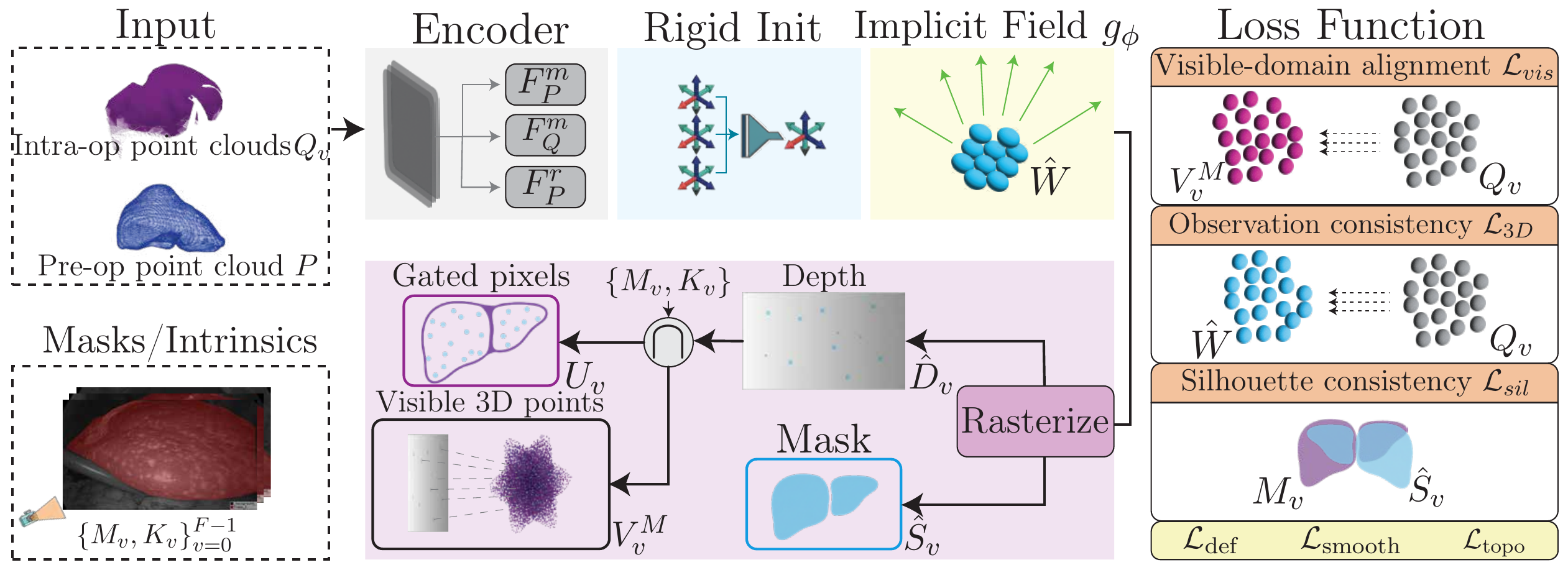}
    \caption{Vis2Reg pipeline. Given preoperative $P$ and intraoperative partial $Q$, the network outputs warped geometry $\hat{W}$ via rigid initialization and non-rigid deformation; differentiable rasterization and mask-gated back-projection construct the visible-domain supervision set $V_v^{M}$ from $(\hat{D}_v,\hat{S}_v)$ and $M_v$.}
    \label{fig1}
\end{figure}

Vis2Reg addresses self-supervised intraoperative liver registration under severe partial visibility and without 3D--2D ground-truth supervision. \rev{The clinical goal is a 3D-to-2D AR overlay, achieved as partial-to-complete 3D registration aligning the complete preoperative cloud $P$ with the partial intraoperative observation $Q$. Each view $Q_v$ is reconstructed from laparoscopic images via monocular depth (DepthAnything~\cite{depthanything}), the liver mask $M_v$, and back-projection with intrinsics $K_v$.}
Given a preoperative liver point cloud $P=\{p_i\}_{i=1}^{N}$ and $F$ view-dependent intraoperative point clouds $\{Q_v\}_{v=0}^{F-1}$ (all in camera coordinates), together with $F$-view masks $\{M_v\}_{v=0}^{F-1}$ and intrinsics $\{K_v\}_{v=0}^{F-1}$, we denote the merged intraoperative observation as $Q \triangleq \bigcup_{v=0}^{F-1} Q_v$. \rev{We use $F{=}3$ as a short \emph{local, non-temporal} window corresponding to a near-static liver; $M_v$ and $K_v$ serve as per-view supervision signals, not model inputs.} The objective is to estimate a rigid alignment and a non-rigid deformation modeled as an implicit displacement field $g_\phi$, producing a warped point cloud $\hat{W}$ that aligns $P$ to the intraoperative observations under visibility-consistent geometric constraints. As shown in Fig.~\ref{fig1}, this is achieved using a geometry-only registration network comprising a robust rigid initialization module and the implicit deformation field $g_\phi$. 
During training, visibility-consistent supervision is constructed from masks via differentiable rasterization technique and mask-gated back-projection, while at inference the model operates using geometric inputs only.

\begin{figure*}[t!]
    \centering
    \includegraphics[width=1\linewidth]{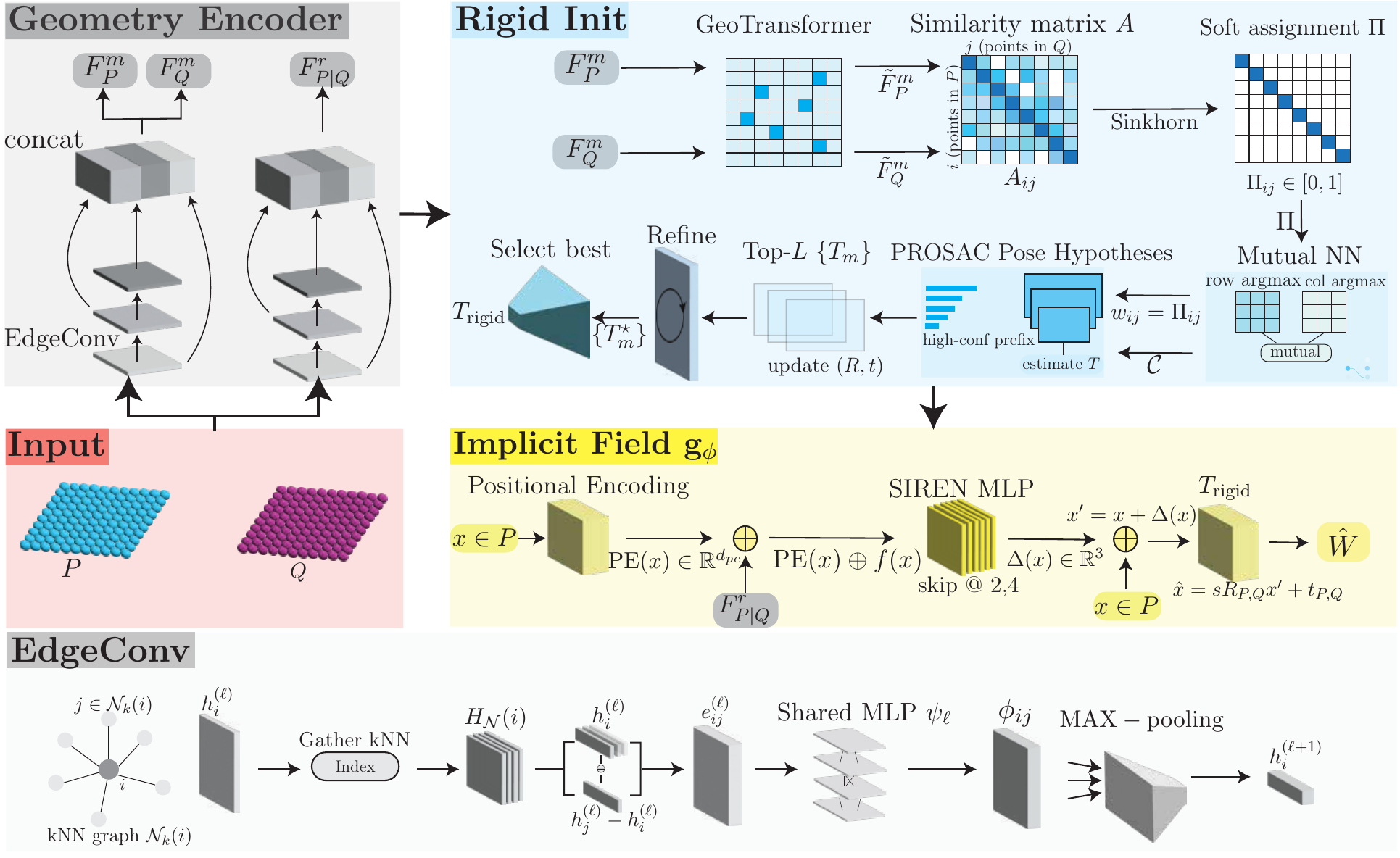}
    \caption{Architecture of the geometry-only registration network in Vis2Reg: EdgeConv feature encoding, rigid initialization ($T_{\mathrm{rigid}}$), and implicit non-rigid deformation field $g_\phi$ producing the warped point cloud $\hat{W}$. Bottom: EdgeConv update block.}
    \label{fig2}
\end{figure*}

\subsection{Geometry-Only Registration Network}
\label{sec:method_network}
As illustrated in Fig.~\ref{fig1}, Vis2Reg predicts the warped geometry $\hat{W}$ via (i) geometric feature encoders, (ii) a robust rigid initialization module for stable global alignment under partial visibility, and (iii) an implicit displacement field $g_\phi$ modeling continuous local deformation. \rev{Concretely, the rigid module builds on a GeoTransformer matcher and the deformation field is implemented as a multilayer perceptron (MLP) with sinusoidal representation network (SIREN) activations, both detailed below.}

\noindent\paragraph{\textbf{Geometric Feature Encoders.}}
In Fig.~\ref{fig2}, we extract multi-scale geometric features using a multi-layer EdgeConv encoder~\cite{wang2019dgcnn}, which captures local shape context through dynamic neighborhood aggregation. Given a point set $X=\{x_i\}_{i=1}^{n}$ with initial features $h_i^{(0)}=x_i$, each layer updates point features via
\begin{equation}
h_i^{(\ell+1)}=\max_{j\in\mathcal{N}_k(i)}\psi_\ell\!\left(\left[h_i^{(\ell)}\,\|\,h_j^{(\ell)}-h_i^{(\ell)}\right]\right),
\end{equation}
where $\psi_\ell(\cdot)$ is a shared MLP and $\mathcal{N}_k(i)$ denotes the $k$-nearest-neighbor ($k$-NN) neighborhood. Concatenating intermediate outputs yields multi-scale features $F(X)$. 
We employ two encoders with distinct roles: a Siamese \emph{matching encoder} $\mathcal{E}_m$ that extracts correspondence features for rigid alignment from $(P,Q)$, while a \emph{registration encoder} $\mathcal{E}_r$ extracts deformation features \rev{conditioned on the pair $(P,Q)$} for the implicit field, i.e., $F_P^{m}=\mathcal{E}_m(P)$, $F_Q^{m}=\mathcal{E}_m(Q)$, and \rev{$F_{P|Q}^{r}=\mathcal{E}_r(P,Q)$, queried at the source points $P$}.

\noindent\textbf{Rigid Seed Initialization.}
To obtain a stable rigid alignment under sparse, noisy observations, we estimate an initial pose via soft correspondence, hypothesis evaluation, and geometric refinement (Fig.~\ref{fig2}). GeoTransformer contextualizes $F_P^m,F_Q^m$ into $\tilde{F}_P^m,\tilde{F}_Q^m$, giving a similarity matrix $A_{ij}=\langle \tilde{f}_i,\tilde{g}_j\rangle/\tau$; Sinkhorn normalization yields a soft assignment $\Pi$ with confidence weights $w_{ij}=\Pi_{ij}$, and mutual nearest neighbors form a correspondence set $\mathcal{C}$. Progressive Sample Consensus (PROSAC) then generates pose hypotheses $\{T_m\}$ from minimal samples in $\mathcal{C}$, ranked by weighted inlier scores; the top hypotheses are refined by trimmed point-to-point and point-to-plane ICP, and the best provides the rigid initialization for $g_\phi$.

\noindent\paragraph{\textbf{Implicit Non-Rigid Deformation Field.}}
We represent non-rigid deformation as a continuous implicit displacement field $g_\phi$ conditioned on geometric features. 
For each source point $x\in P$, positional encoding $\mathrm{PE}(x)\in\mathbb{R}^{d_{pe}}$ is concatenated with its \rev{pair-conditioned} registration feature \rev{$f(x)=F_{P|Q}^r(x)$} to form the field input $z(x)=\mathrm{PE}(x)\oplus f(x)$. A SIREN-based MLP models the implicit field and predicts a displacement
$\Delta(x)=g_\phi(z(x))\in\mathbb{R}^3$,
which produces a locally deformed point
$x'=x+\Delta(x)$.
The final warped geometry is obtained by applying the rigid alignment to the deformed point:
\begin{equation}
\hat{x}=sR_{P,Q}\, x'+t_{P,Q},\qquad \hat{W}=\{\hat{x}\mid x\in P\}.
\end{equation}
The resulting warped point cloud $\hat{W}$ is then used for visibility-aware supervision. \rev{Since both the pair-conditioned feature $F_{P|Q}^r$ and the rigid transform $(sR_{P,Q},t_{P,Q})$ depend on $Q$, the deformation adapts to each intraoperative input rather than being a fixed function of $P$.}

\subsection{Visibility-Aware Self-Supervised Learning}
\label{sec:loss}
The observed clouds $\{Q_v\}_{v=0}^{F-1}$ are view-dependent visible subsets of the surface, with no 3D registration ground truth. Vis2Reg therefore constructs a \emph{mask-consistent visible-domain 3D supervision signal}, restricting supervision to geometrically observable regions defined by the masks and guiding deformation only by physically valid observations. \rev{Unlike the symmetric rendered-mask consistency of Self-P2IR~\cite{zhou_landmark-free_2025}, our supervision is mask-gated and one-way (observation-to-model), so unobserved model regions are never penalized.}

\noindent\paragraph{\textbf{Visible-domain construction.}}
Let $\Omega$ denote the discrete $W\times H$ image lattice. From the warped cloud $\hat{W}$, intrinsics $\{K_v\}$, and masks $\{M_v\}$, we render depth and silhouette per view by differentiable point rasterization:
\begin{equation}
(\hat{S}_v, \hat{D}_v)=\mathcal{R}_{K_v}(\hat{W}), \quad v=0,\ldots,F-1,
\end{equation}
where $\mathcal{R}_{K_v}$ is parameterized by $K_v$ and $\hat{D}_v(u)=0$ marks empty pixels. We define the mask-consistent visible pixel set $U_v=\{u\in\Omega \mid \hat{D}_v(u)>0 \wedge M_v(u)=1\}$ and back-project it to an explicit visible-domain 3D supervision set:
\begin{equation}
V^{M}_v=
\left\{
\pi^{-1}_{K_v}\!\big(u,\hat{D}_v(u)\big)
\;\middle|\;
u\in U_v
\right\},
\end{equation}
where $\pi^{-1}_{K_v}$ is the analytic back-projection from a pixel-depth pair to a camera-frame 3D point.

\noindent\paragraph{\textbf{Self-Supervised Objectives.}}
We optimize a weighted combination of visibility-aware alignment losses and deformation regularization:
\begin{equation}
\mathcal{L}=\lambda_{3D}\mathcal{L}_{3D}+\lambda_{vis}\mathcal{L}_{vis}+\lambda_{sil}\mathcal{L}_{sil}
+\lambda_{def}\mathcal{L}_{def}+\lambda_{smooth}\mathcal{L}_{smooth}+\lambda_{topo}\mathcal{L}_{topo}.
\end{equation}
We denote the one-way Chamfer as $\mathrm{CD}_{\rightarrow}(A,B)=\frac{1}{|A|}\sum_{a\in A}\min_{b\in B}\|a-b\|_2^2$ and the symmetric Chamfer as $\mathrm{CD}(A,B)=\mathrm{CD}_{\rightarrow}(A,B)+\mathrm{CD}_{\rightarrow}(B,A)$.

The visibility-aware alignment terms are an observation-to-model one-way Chamfer $\mathcal{L}_{3D}=\frac{1}{F}\sum_{v}\mathrm{CD}_{\rightarrow}(Q_v,\hat{W})$ (handling the partial visibility of $Q_v$), a symmetric visible-domain term $\mathcal{L}_{vis}=\frac{1}{F}\sum_{v}\mathrm{CD}(V_v^{M},Q_v)$ aligning the mask-gated set $V_v^{M}$ with $Q_v$, and a silhouette term $\mathcal{L}_{sil}=\frac{1}{F}\sum_{v}\big[\mathrm{BCE}(\hat{S}_v,M_v)+\mathrm{DiceLoss}(\hat{S}_v,M_v)\big]$.
\rev{ The displacement field is regularized by deformation magnitude ($\mathcal{L}_{def} =\tfrac{1}{N}\sum_{i}\|\Delta_i\|_2^2$), local smoothness ($\mathcal{L}_{smooth}=\tfrac{1}{N}\allowbreak\sum_{i}\allowbreak\tfrac{1}{|\mathcal{N}(i)|}\allowbreak\sum_{j\in\mathcal{N}(i)}\allowbreak\|\Delta_i-\Delta_j\|_2^2$), and local distance or topology preservation ($\mathcal{L}_{topo}=\tfrac{1}{N}\allowbreak\sum_{i}\allowbreak\tfrac{1}{|\mathcal{N}(i)|}\allowbreak\sum_{j\in\mathcal{N}(i)}\allowbreak\big(\|\hat{x}_i-\hat{x}_j\|_2\allowbreak-\|x_i-x_j\|_2\big)^2$).}
\rev{\noindent{Here,} $\mathcal{N}(i)$ is the $k$-NN neighborhood of $x_i$ and $\Delta_i=\Delta(x_i)$.}

\noindent\paragraph{\textbf{Training schedule.}} We adopt a two-stage synthetic$\rightarrow$real schedule: the matching and rigid-seed modules are pretrained on synthetic data with ground-truth (GT) poses (non-rigid field disabled), then trained on real data (no GT) with a short rigid warm-up followed by joint visibility-aware optimization.

\section{Experiments}
\subsection{Dataset and Implementation Details}
We evaluate Vis2Reg on the public P2I-LReg dataset, a real intraoperative liver registration benchmark with 346 keyframes from 21 patients~\cite{zhou_landmark-free_2025}. Each sample provides a preoperative 3D liver model, an intraoperative sparse point cloud, a binary segmentation mask, and camera intrinsics. Following the official patient-level protocol, patients are split into 12/4/5 for training/validation/testing, and results are averaged over 5-fold cross-validation. For robust rigid initialization, we additionally use the patient-specific Landmark-Free Synthetic Dataset (red-green-blue plus depth (RGB-D) observations rendered from preoperative models via physics-based Blender simulation with diverse viewpoints and occlusions; 2500 samples per patient; 60\%, 20\%, and 20\% split; training-fold patients only).

Inputs are $P$, $Q$, and multi-view masks and intrinsics $\{M_v, K_v\}$, all in camera coordinates at real-world scale (no centering and normalization). During training we apply random dropout and jittering to $P$ and statistical denoising to $Q$; both are resampled or zero-padded to $N_{\max}=6000$.

We use AdamW ($\mathrm{lr}=3\times10^{-4}$, $\mathrm{wd}=10^{-4}$) with cosine scheduling and automatic mixed precision (AMP). In Stage-2 real-data training, we train for 60 epochs (batch size 1) with a 20-epoch rigid warm-up (non-rigid branch and deformation regularizers frozen), followed by joint optimization; a 3-frame multi-view input is used by default. Loss weights are $\lambda_{3D}=0.5$, $\lambda_{vis}=1.2$, $\lambda_{sil}=1.0$ (BCE and Dice weights both 1.0), $\lambda_{def}=0.1$, $\lambda_{smooth}=0.1$, and $\lambda_{topo}=0.3$. For differentiable rasterization, \textit{points-per-pixel}=16 and the maximum number of visible points is 5000. On synthetic data, we report relative rotation error (RRE) and relative translation error (RTE), defined as $\mathrm{RRE}=\arccos((\mathrm{tr}(\mathbf{R}_{\mathrm{pred}}^\top \mathbf{R}_{\mathrm{gt}})-1)/2)$ (degrees) and $\mathrm{RTE}=\|\mathbf{t}_{\mathrm{pred}}-\mathbf{t}_{\mathrm{gt}}\|_2$ (mm). On real data, we report Chamfer Distance ($\mathcal{L}_{3D}$) and silhouette Dice. \rev{Dice (our primary AR-overlay metric) measures overlap between the rasterized registered silhouette and the liver mask, whereas CD is a complementary surface-distance measure, not a target registration error. All baselines use the same patient-level folds and protocol} implemented in PyTorch on NVIDIA L40S GPU.
\begin{table}[!t]
\centering
\caption{Comparison on synthetic rigid initialization and real intraoperative registration (mean $\pm$ std). RRE and RTE are reported only for rigid-comparable methods; non-rigid methods are marked as ``--'' in the synthetic block.}
\label{tab:cmp}

{\fontsize{8}{9}\selectfont
\setlength{\tabcolsep}{3.2pt}
\renewcommand{\arraystretch}{0.9}

\begin{tabularx}{\linewidth}{@{}>{\raggedright\arraybackslash}X c c @{\hspace{3pt}} c c@{}}
\toprule
\multirow{2}{*}{Method} & \multicolumn{2}{c}{Synthetic (Rigid Init)} & \multicolumn{2}{c}{Real Intraoperative (Final)} \\
\cmidrule(lr){2-3}\cmidrule(l){4-5}
& RRE ($^\circ$)$\downarrow$ & RTE (mm)$\downarrow$ & Dice (\%)$\uparrow$ & CD (mm)$\downarrow$ \\
\midrule
GeoTransformer~\cite{qin_geotransformer_2023}
& 31.30 & 78.10 & 71.16 $\pm$ 6.46 & 3.43 $\pm$ \textbf{0.96} \\

DPF~\cite{prokudin_dynamic_2023}
& -- & -- & 72.19 $\pm$ \underline{5.97} & 3.31 $\pm$ 1.18 \\

PointSetReg~\cite{zhao_correspondence-free_2024}
& -- & -- & 75.24 $\pm$ \textbf{5.92} & 3.20 $\pm$ \underline{1.03} \\

Self-P2IR~\cite{zhou_landmark-free_2025}
& 0.21 & 1.32 & \underline{78.89} $\pm$ 6.76 & \underline{2.97} $\pm$ 1.06 \\

\textbf{Ours (Vis2Reg)}
& \textbf{0.08} & \textbf{0.26} & \textbf{92.60} $\pm$ 7.24 & \textbf{1.43} $\pm$ 1.26 \\
\bottomrule
\end{tabularx}
}
\end{table}
 
\subsection{Comparison and Ablation Study}
On the synthetic benchmark (with 3D GT), Vis2Reg achieves the best rigid initialization performance in the synthetic block of Table~\ref{tab:cmp}, with RRE $0.08^\circ$ and RTE $0.26$ mm, outperforming GeoTransformer (RRE $31.30^\circ$, RTE $78.10$ mm) and Self-P2IR (RRE $0.21^\circ$, RTE $1.32$ mm), indicating a more reliable rigid seed under low overlap. DPF and PointSetReg are non-rigid models and are therefore not included in the synthetic rigid comparison. \rev{The GeoTransformer entry uses the standalone matcher with its built-in closed-form pose estimation, with no robust outlier rejection (e.g., RANSAC) and no ICP refinement, hence the higher RRE/RTE (Table~\ref{tab:cmp}). Vis2Reg instead uses the full pipeline (GeoTransformer matching, mutual filtering, PROSAC, trimmed ICP), reconciling the gap with the robust-estimator values reported on the source dataset~\cite{zhou_landmark-free_2025}.}
On the real intraoperative benchmark (Table~\ref{tab:cmp}), Vis2Reg achieves the best mean performance in both final registration metrics, with Dice $92.60\pm7.24$\% and CD $1.43\pm1.26$ mm. Compared with the strongest prior method (Self-P2IR), Vis2Reg improves Dice by 13.71 pp (percentage points) and reduces CD by 1.54 mm. These gains are consistent with our visibility-aware design, where mask-gated visible-domain supervision constrains optimization to observed regions and mitigates drift under severe occlusion and partial visibility. \rev{Importantly, Self-P2IR already attains a near-perfect rigid init (RRE $0.21^\circ$, RTE $1.32$ mm), comparable to ours, yet trails Vis2Reg by $13.71$ pp in Dice; as both start from a strong rigid seed, this gain stems from the visibility-aware non-rigid supervision, not initialization. The ablations (Table~\ref{tab:ablation}) corroborate this: removing visibility-aware supervision ($\mathcal{L}_{vis}$ or mask-gating) alone reduces Dice by $13$--$18$ pp.}

\begin{figure}[t]
    \centering
    \includegraphics[width=1\linewidth,trim=8 10 8 10,clip]{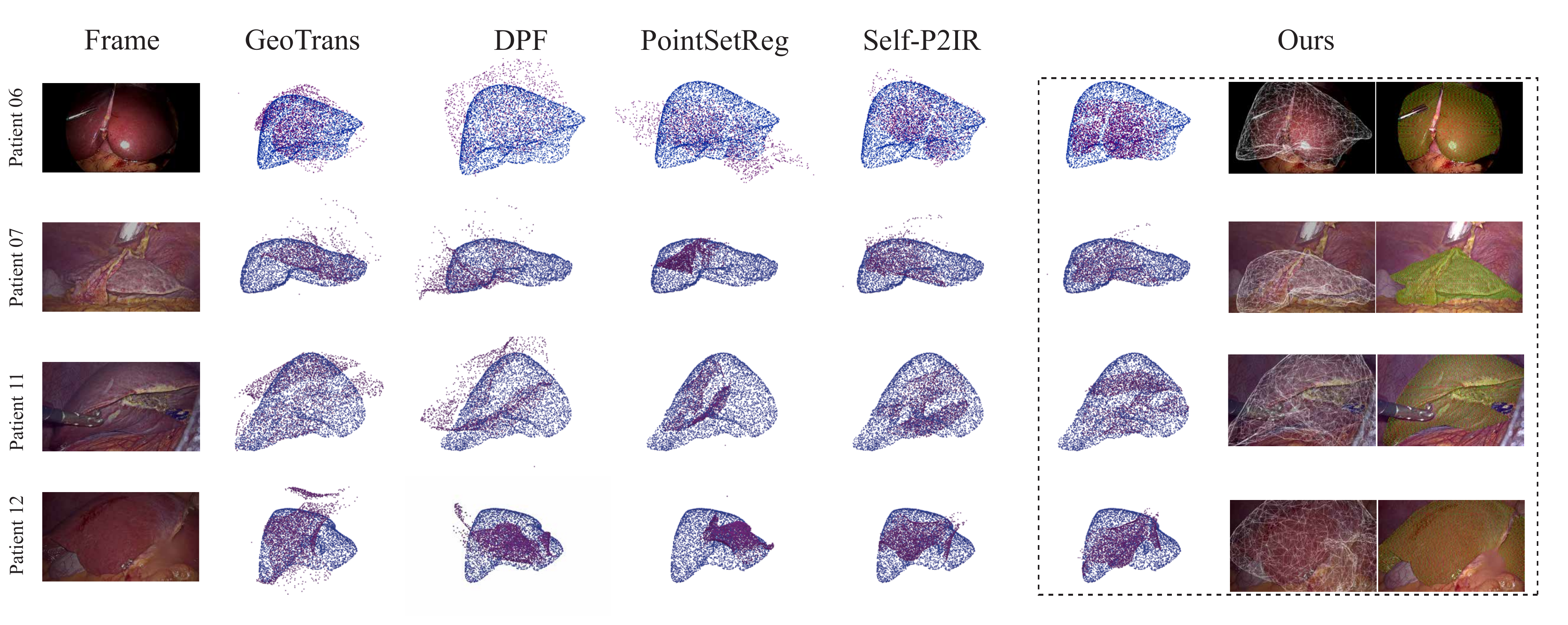}
    \caption{Qualitative comparison on intraoperative keyframes. Col.~1: input. Cols.~2--5: baselines. Col.~6: Vis2Reg point cloud. Col.~7: Vis2Reg mesh overlay. Col.~8: visible-region fitting. Blue: preoperative; purple: intraoperative.}
    \label{fig3}
\end{figure}

Fig.~\ref{fig3} shows qualitative results on representative keyframes: Vis2Reg achieves tighter preoperative--intraoperative alignment than the baselines and better adapts the visible surface to observations while preserving globally plausible anatomy. Vis2Reg runs at 111.38 ms per forward pass with 1.95 GB GPU memory on an NVIDIA L40S, supporting near-real-time intraoperative use. \rev{The reported inference time covers forward registration after $Q_v$ and $M_v$ are available, excluding depth and mask reconstruction.}
\begin{table}[!t]
\centering
\caption{Ablations on real data. ``With gate'' refers to $U_v=\{\hat{D}_v>0 \land M_v=1\}$, while ``without gate'' uses $U_v=\{\hat{D}_v>0\}$.}
\label{tab:ablation}

{\fontsize{8}{9}\selectfont
\setlength{\tabcolsep}{3.2pt}
\renewcommand{\arraystretch}{0.9}

\begin{tabularx}{\linewidth}{@{}>{\raggedright\arraybackslash}X c c c c c c@{}}
\toprule
Variant & $\mathcal{L}_{vis}$ & gate & 1-way $\mathcal{L}_{3D}$ & seed & Dice$\uparrow$ & CD$\downarrow$ (mm) \\
\midrule
w/o $\mathcal{L}_{vis}$ & N & Y & Y & Y & 74.29 & 3.12 \\
w/o mask-gating         & Y & N & Y & Y & 79.57 & 3.04 \\
sym.\ $\mathcal{L}_{3D}$ & Y & Y & N & Y & \underline{84.48} & \underline{2.98} \\
weak rigid init         & Y & Y & Y & N & 69.32 & 3.64 \\
\addlinespace[1pt]
\textbf{Full (Vis2Reg)} & \textbf{Y} & \textbf{Y} & \textbf{Y} & \textbf{Y} & \textbf{92.60} & \textbf{1.43} \\
\bottomrule
\end{tabularx}
}
\end{table}

Ablation results in Table~\ref{tab:ablation} validate each design choice: removing $\mathcal{L}_{vis}$ or mask-gating, replacing one-way $\mathcal{L}_{3D}$ with symmetric Chamfer, and weakening rigid initialization each degrade performance---the last causing the largest drop---supporting mask-gated visible-domain supervision, the observation-to-model formulation, and a strong rigid seed.

\section{Conclusion}
Vis2Reg is a visibility-aware, landmark-free, geometry-only framework for intraoperative liver 3D--2D registration under severe occlusion, coupling robust rigid initialization with an implicit non-rigid deformation field and differentiable point rasterization for mask- and camera-based self-supervision without 3D ground truth. On real intraoperative data, it significantly improves Dice and Chamfer Distance over prior methods while retaining near-real-time performance. \rev{We restrict our evaluation and claims to the P2I-LReg benchmark. Evaluating tumour and vascular structures would require internal anatomical ground truth~\cite{rabbani2022methodology}, unavailable here, and is left to future work.}

\begin{credits}
\subsubsection{\ackname}
The project was funded by the Engineering and Physical Sciences Research Council [Grant No. UKRI914].

\subsubsection{\discintname}
The authors have no competing interests to declare that are relevant to the content of this article.
\end{credits}

%
%
\bibliographystyle{splncs04}
\bibliography{references}  

\end{document}